\documentclass[11pt,a4paper]{article}
\usepackage[hyperref]{emnlp2018}
\usepackage{times}
\usepackage{latexsym}
\usepackage{sidecap}
\usepackage{paralist} 

\usepackage{algorithm}
\usepackage[noend]{algpseudocode}
\usepackage{graphicx}
\usepackage{amsmath}
\graphicspath{ {images/} }
\usepackage{pgf}
\usepackage{caption}
\usepackage{microtype}
\usepackage{pifont}

\usepackage{url}

\aclfinalcopy % Uncomment this line for the final submission
\def\aclpaperid{1351} %  Enter the acl Paper ID here

\newcommand{\todom}[1]{}
\newcommand{\todon}[1]{}
\newcommand{\todoa}[1]{}
\newcommand{\todok}[1]{}

\title{Accelerating Asynchronous Stochastic Gradient Descent for Neural Machine Translation}

\author{Nikolay Bogoychev\textsuperscript{\ding{76}} Marcin Junczys-Dowmunt\textsuperscript{\ding{40}} Kenneth Heafield\textsuperscript{\ding{76}} Alham Fikri Aji\textsuperscript{\ding{76}}\\
\ding{76} School of Informatics, University of Edinburgh, Edinburgh, United Kingdom\\
\ding{40} Microsoft, 1 Microsoft Way, Redmond, WA 98121, USA \\
{\tt \{n.bogoych, a.fikri, kheafiel\}@ed.ac.uk, marcinjd@microsoft.com} \\}
\date{}

\begin{document}
\maketitle
\begin{abstract}
In order to extract the best possible performance from asynchronous stochastic gradient descent one must increase the mini-batch size and scale the learning rate accordingly. In order to achieve further speedup we introduce a technique that delays gradient updates  effectively increasing the mini-batch size. Unfortunately with the increase of mini-batch size we worsen the stale gradient problem in asynchronous stochastic gradient descent (SGD) which makes the model convergence poor. We introduce local optimizers which mitigate the stale gradient problem and together with fine tuning our momentum we are able to train a shallow machine translation system 27\% faster than an optimized baseline with negligible penalty in BLEU. %Further using our methods we are able to train a deep NMT system 23\% than the baseline and reach better model cross-entropy.
\end{abstract}

\section{Introduction}

With training times measured in days, parallelizing stochastic gradient descent (SGD) is valuable for making experimental progress and scaling data sizes.  Synchronous SGD sums gradients computed by multiple GPUs into one update, equivalent to a larger batch size.  But GPUs sit idle unless workloads are balanced, which is difficult in machine translation and other natural language tasks because sentences have different lengths.  Asynchronous SGD avoids waiting, which is faster in terms of words processed per second.  However  asynchronous SGD suffers from stale gradients \cite{tf_synchr} that degrade convergence, resulting in an almost no improvement in time to convergence \cite{omnivore}. This paper makes asynchronous SGD even faster and deploys a series of convergence optimizations.
 
In order to achieve fastest training (and inspired by \newcite{fb_imagenet} we increase the mini-batch size, making the matrix operations more efficient and reducing the frequency of gradient communication for the optimizer step. Unlike their task (image classification), text training consumes a lot of GPU Memory (Table~\ref{mini_batch_memory}) for word embedding activations making it impossible to fit mini-batches of similar magnitude as \citet{fb_imagenet}.  

Our main contributions are as follows:
\begin{compactenum}
\item We introduce a delayed gradient updates which allow us to work with much larger mini-batches which would otherwise not be possible due to limited GPU memory.
\item We introduce local optimizers which run on each worker to mitigate the extra staleness and convergence issues \cite{largembconvergence,Keskar2017} caused by large mini-batches.
\item We highlight the importance of tuning the optimizer momentum and show how it can be used as a cooldown strategy.
\end{compactenum}
%Therefore, we emulate larger batches by running $\tau$ batches locally and summing them on each GPU.  As Section~\ref{largebatch} confirms, larger batches increase speed (Table \ref{mini_batch_memory}) but also make convergence per epoch slower \cite{largembconvergence,Keskar2017}, which we mitigate by using local optimizers on each GPU and tuning our momentum. While learning rate customization receives much attention in the literature, the momentum parameters of the Adam optimizer \cite{adam} are relatively neglected.  We add momentum cooldown and tuning to improve convergence.

\begin{table}[ht]
\centering
\begin{tabular}{@{}l@{\ \ }r@{\ \ }r@{\ \ }r@{\ \ }r@{\ \ }r@{}}
\textbf{VRAM}& $\tau$ &\textbf{Words} &\textbf{WPS}\\
3 GB     & 1  & 3080      & 19.5k \\
7 GB     & 1  & 7310      & 36.6k \\
10 GB    & 1  & 10448     & 40.2k \\
20* GB   & 2  & 20897     & 44.2k \\
30* GB   & 2  & 31345     & 46.0k \\
40* GB   & 4  & 41794     & 47.6k
\end{tabular}
\caption{Relationship between the GPU Memory (VRAM) budget for batches (* means emulated by summing $\tau$ smaller batches), number of source words processed in each batch and words-per-second (WPS) measured on a shallow model.}
\label{mini_batch_memory}
\end{table}

\section{Experiments}
\label{experiments}
This section introduces each optimization along with an intrinsic experiment on the WMT 2016 Romanian$\rightarrow$English task \cite{wmt16-task}.

The translation system is equivalent to \citet{wmt16}, which was the first place constrained system (and tied for first overall in the WMT16 shared task.).  The model is a shallow bidirectional GRU \cite{gru} encoder-decoder trained on 2.6 million parallel sentences. Due to variable-length sentences, machine translation systems commonly fix a memory budget then pack as many sentences as possible into a dynamically-sized batch. The memory allowance for mini-batches in our system is 3 GB (for an average batch size of 2633 words). Adam \cite{adam} is used to perform asynchronous SGD with learning rate of 0.0001. This is our baseline system. We also compare with a synchronous baseline which uses modified Adam parameters, warmup of 16000 mini-batches and inverse square root cooldown following \citet{google_att}. We used 4 Tesla P 100 GPUs in a single node with the Marian NMT framework for training \cite{mariannmt}. Since we apply optimizations over asynchronous SGD we performed a learning rate and mini-batch-size parameter sweep over the baseline system and settled on a learning rate of 0.00045 and 10 GB memory allowance for mini-batches (average batch size of 10449 words). This is the fastest system we could train without sacrificing performance before adding our improvements. In our experiments on Table \ref{roen_results} we refer to this system as "Optimized asynchronous". All systems were trained until 5 consecutive stalls in the cross-entropy metric of the validation set. Note that some systems require more epochs to reach this criteria which indicates poor model convergence.

\subsection{Larger Batches and delayed updates}
\label{largebatch}
This experiment aims to increase speed, in words-per-second (WPS), by increasing the batch size.  Larger batches have two well-known impacts on speed: making more use of GPU parallelism and communicating less often.  

After raising the batch size to the maximum that fits on the GPU,\footnote{The Tesla P100 has 16 GB of GPU memory and we opt to use 10 GBs of mini-batches and the rest is used to store model parameters, shards, optimizers and additional system specific elements such as the cache vectors for gradient dropping \cite{drop}.} we emulate even larger batches by processing multiple mini-batches and summing their gradients locally without sending them to the optimizer.  This still increases speed because communication is reduced (Table~\ref{mini_batch_memory}).  We introduce parameter $\tau$, which is the number of iterations a GPU performs locally before communicating externally as if it had run one large batch.  The Words-per-second (WPS) column on Table~\ref{mini_batch_memory} shows the effect on corpora processing speed when applying delayed gradients updates for different values of $\tau$. While we reduce the overall training time if we just apply delayed gradient updates we worsen the overall convergence (Table \ref{roen_results}).

When increasing the mini-batch size $\tau$ times without touching the learning rate, we effectively do $\tau$ times less updates per epoch. On the surface, it might seem that these less frequent updates are counterbalanced by the fact that each update is accumulated over a larger batch. But practical optimization heuristics like gradient clipping mean that in effect we end up updating the model less often, resulting in slower convergence. \citet{fb_imagenet} recommend scaling the learning rate linearly with the mini-batch size in order to maintain convergence speed.

\subsection{Warmup}
\citet{fb_imagenet} point out that just increasing the learning rate performs poorly for very large batch sizes, because when the model is initialized at a random point, the training error is large. Large error and large learning rate result in bad "jerky" updates to the model and it can't recover from those. \citet{fb_imagenet} suggest that initially model updates should be small so that the model will not be pushed in a suboptimal state. Afterwards we no longer need to be so careful with our updates.

\subsubsection{Lowering initial learning rate}
\citet{fb_imagenet} lower the initial learning rate and gradually increase it over a number of mini-batches until it reaches a predefined maximum. This technique is also adopted in the work of \citet{google_att}. This is the canonical way to perform warmup for neural network training.

\subsubsection{Local optimizers}
We propose an alternative warm up strategy and compare it with the canonical method.
Since we emulate large batches by running multiple smaller batches, it makes sense to consider whether to optimize locally between each batch by adapting the concept of local per-worker optimizers from \citet{elasticSGD}. 
In asynchronous SGD setting each GPU has a full copy of the model as well as the master copy of $1/N$th of the parameters in its capacity as parameter server. We use the local optimizers to update the local model shard in between delayed gradient updates, which helps mitigate staleness. Unlike prior work, we also update the shard of the global model that happens to be on the same GPU.  Local updates are almost free because we avoid remote device communication.% and the additional computation time that the local optimizer carries is negligible.

Updating the parameter shard of the global model bears some resemblance to the Hogwild method \cite{hogwild} as we don't synchronize the updates to the shard, however, global updates are still synchronised. As before, once every $\tau$ iterations we run a global optimizer that updates the sharded parameter set and then distributes the updated model across all devices. Any local model divergences are lost at this point. We found that this strategy improves model convergence in the early epochs but tends to be harmful later on. We hypothesize that initially partial model updates reduce staleness, but when the model starts to converge, local optimizers introduce extra noise in the training, which is harmful. We use local optimizers purely as a warmup strategy, turning them off after the initial phase of the training. Empirically, we found that we can get the best convergence by using them for the first 4000 mini-batches that each device sees. On Table \ref{roen_results} we compare and contrast the two warmup strategies. By itself learning-rate warmup offers slower convergence but to a better point compared to local optimizers. The reader may notice that if we apply delayed gradient updates, the effective batch size that the global optimizer deals with is $\tau$ times larger than the mini-batch size on which the local optimizers runs. Therefore we use $\tau$ times lower learning rate for the local optimizers compared to the global optimizers.

\begin{table*}[ht]
\centering
\begin{tabular}{llllll}
\textbf{System}              & \textbf{Time (hours)}  & \textbf{Epochs} & \textbf{BLEU} & \textbf{CE}   & \textbf{WPS}\\
synchronous                  & 14.3                   & 11              & 35.3          & 50.63         & 15.7k       \\
asynchronous (0)             & 12.2                   & 13              & 35.61         & 50.47         & 19.5k       \\
(0) + \citet{drop}           & 6.23                   & 12              & 35.16         & 50.86         & 26.9k       \\
optimized asynchronous (1)   & 4.97                   & 10              & 35.56         & 50.90         & 40.2k       \\
(1) + \citet{drop}           & 4.32                   & 11              & 35.16         & 52.02         & 41.5k       \\
(1) + delayed updates $\tau=2$ (2) & 4.20             & 11              & 34.82         & 51.68         & 44.2k       \\
(2) + local optimizers (3)   & 3.66                   & 10              & 35.45         & 51.32         & 44.2k       \\
\textbf{(3) + momentum tuning (4)}&\textbf{3.66}      & \textbf{10}     &\textbf{35.48} &\textbf{50.87} &\textbf{44.2k}\\
(2) + warmup (5)             & 4.87                   & 13              & 35.29         & 50.78         & 44.2k       \\ 
(5) + momentum tuning (6)    & 3.98                   & 11              & 35.76         & 50.73         & 44.2k
\end{tabular}
\caption{Romanian-English results from our exploration and optimizations. We also compare our methods against the work of \citet{drop} which also reduces communication. We use system (1) as our reference baseline upon which we improve. The system that achieved the best training time is bolded.}
\label{roen_results}
\end{table*}

\subsection{Momentum cooldown and tuning}
\citet{fb_imagenet} and \citet{google_att} both employ cooldown strategies that lower the learning rate towards the end of training. Inspired by the work of \citet{omnivore} however we decided to pursue a different cooldown strategy by modifying the momentum inside Adam's parameters.

%\todok{Here to make it compile by moving citation.  Lots of text to make it compile!}
Momentum tuning is not a well explored area in deep learning. Most researchers simply use the default values for momentum for a chosen optimizer \cite{omnivore} (in the case of NMT, this is usually Adam). \citet{omnivore} argue that this is an oversight especially when it comes to asynchronous SGD, because the asynchronisity adds extra implicit momentum to the training which is not accounted for. Because of this, asynchronous SGD has been deemed ineffective, as without momentum tuning, the observed increase in training speed is negated by the lower convergence rate, resulting in near-zero net gain \cite{tf_synchr}. However, \citet{omnivore} show that after performing a grid search over momentum values, it is possible to achieve convergence rates typical for synchronous SGD even when working with many asynchronous workers. The downside of momentum tuning is that we can't offer rule-of-thumb values, as they are individually dependent on the optimizer used, the neural model, the number of workers and the batch size. In our experiments, we lowered the overall momentum and in addition performed momentum cooldown where we reduced the momentum of our optimizer (Adam) after the first few thousand batches.

%Another place where non-default Adam settings are used is in the work by \citet{google_att} however they do not elaborate on it. Empirically we found that it is not possible for the Transformer model to converge without modifying the momentum values. 

\subsection{Results}
\begin{figure}[ht]
\scalebox{0.53}{\input{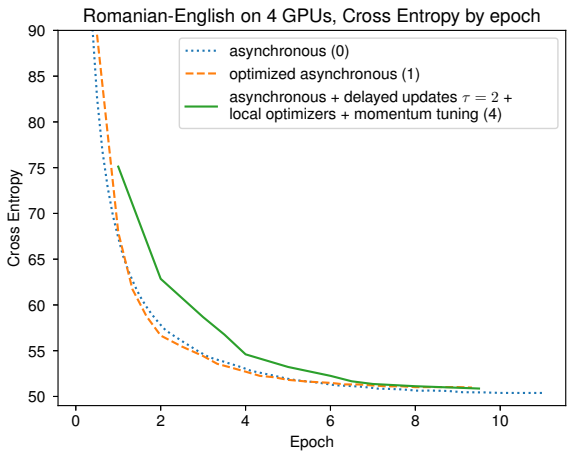}}
    \caption{Cross-entropy training progression per epoch for our ro-en systems.}
    \label{ce_ro}
\end{figure}

\begin{figure}[ht]
\scalebox{0.53}{\input{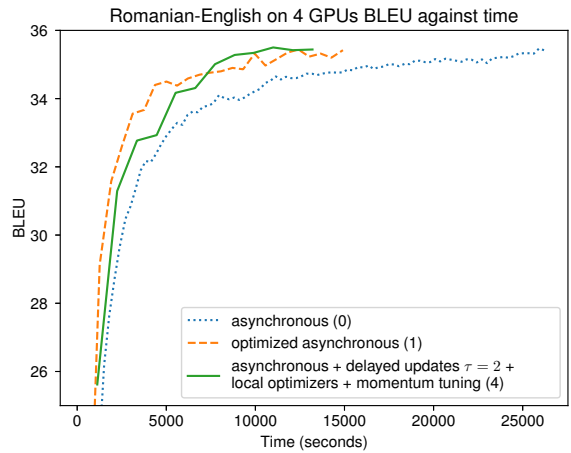}}
    \caption{BLEU scores for our ro-en systems. }
    %Due to the higher training speed and similar per-epoch training progression, our system achieves the target blue score faster. }
    \label{bleu_ro}
\end{figure}

Table \ref{roen_results} shows the effect of modifying momentum values. When using just delayed gradient updates, training is noticeably faster, but there are significant regressions in BLEU and CE (system 2). In order to mitigate those, when using delayed gradient updates, we tune the momentum and apply momentum cooldown on top of either of our warmup strategies. By doing this we not only further reduce training time, but also recover the loss of accuracy. Compared to the optimized baseline system (1), our best system (4) reduces the training time by 27\%. Progression of the training can be seen on figures \ref{ce_ro} and \ref{bleu_ro}. Our system starts poorly compared to the baselines in terms of epoch-for-epoch convergence, but catches up in the later epochs. Due to faster training speed however, the desired BLEU score is achieved faster (Figure \ref{bleu_ro}).

Local optimizers as a warmup strategy show faster convergence compared to learning rate warmup at almost no penalty to BLEU or cross-entropy (System 4 vs system 6). Against the system used in WMT 16 \cite{wmt16}, we achieve nearly 4 times faster training time with no discernible penalty in BLEU or CE. In contrast, the other communication reducing method tested, the work of \citet{drop}, is slower than our work and achieves worse BLEU and CE.

\subsection{Using even larger mini-batches}
\begin{figure}[ht]
\scalebox{0.53}{\input{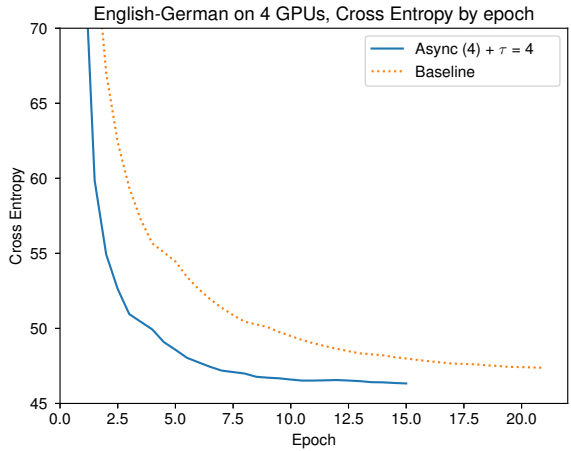}}
    \caption{CE scores for our en-de systems.}
    \label{ce_de}
\end{figure}

\begin{figure}[ht]
\scalebox{0.53}{\input{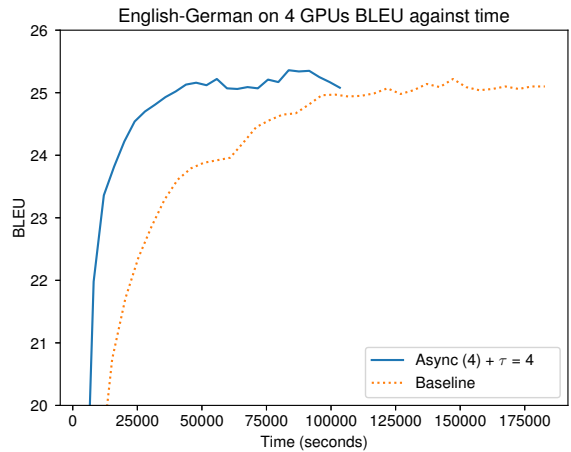}}
    \caption{BLEU scores for our en-de systems.}
    \label{bleu_de}
\end{figure}

We can achieve even greater processing speed by further increasing $\tau$ but we were unable to maintain the same convergence with the Romanian-English shallow model. We found that larger $\tau$ values are useful when dealing with the larger deep RNN models. With deep RNN models the parameters take the majority of the available VRAM leaving very little for mini-batches. In this scenario we can apply $\tau=4$ without negative effect towards convergence. We demonstrate the effectiveness of larger $\tau$ on Table \ref{ende_results}. The baseline system is equivalent to the winning system for English-German at the WMT 2017 competition \cite{wmt17}. The baseline is trained with synchronous SGD and our system uses asynchronous SGD, delayed gradient updates by a factor of 4, local optimizers and the momentum is tuned and further reduced after the first 16000 mini-batches. We found learning rate of 0.0007 to work the best. We do not report the numbers for asynchronous baseline because we were unable to achieve competitive BLEU scores without using delayed gradient updates. We speculate this is because with this type of deep model, our mini-batch size is very small leading to very jerky and unstable training updates. Larger mini-batches ensure the gradients produced by different workers are going to be closer to one another. Our training progression can be seen on figures \ref{ce_de} and \ref{bleu_de}. We show that even though we use 4 times larger mini-batches we actually manage to get lower Cross-Entropy epoch for epoch compared to the baseline (Figure \ref{ce_de}). This coupled with out higher training speed makes our method reach the best BLEU score 1.6 times faster than the baseline (Figure \ref{bleu_de}).
%actually lr is 0.0006776 but that is harder to explain
\begin{table}[ht]
\centering
\begin{tabular}{llllll}
\textbf{System}              & \textbf{Time (h)}  & \textbf{BLEU} & \textbf{CE} \\
Baseline                     & 51.3                   & 25.1          & 47.31    \\
Async (4) + $\tau=4$         & 39.7                   & 25.07         & 46.59 

\end{tabular}
\caption{Training times for English-German deep RNN system trained on WMT17 data. Our asynchronous system includes the optimizations of system (4) from Table \ref{roen_results}.}
\label{ende_results}
\end{table}

\section{Related work}
We use larger mini-batches and delay gradient updates in order to increase the speed at which the dataset is processed. The principal reason why this works is because when mini-batch size is increased $n$ (also includes delayed updates) times, communication is reduced by the same amount. This aspect of our work is similar to the work of \citet{drop} where they drop the lower 99\% of the gradient updates based on absolute value thus reducing the memory traffic. Compared with them we achieve faster dataset processing speed and also better model convergence as shown on Table \ref{roen_results}.

Independently from us \citet{deep_gradient_compression} extend the work of \citet{drop} aiming to  reduce gradient communication without suffering any of the negative effects we have noted. In process they independently arrive to some of the methods that we use, notably tuning the momentum and applying warmup to achieve better convergence.

Independently from us \citet{adafactor} have done further exploratory work on ADAM's momentum parameters using the Transformer model \cite{google_att} as a case study and have offered a mathematical explanation about why different stages of the training require different momentum values.\footnote{This work was published on 11.04.2018.}

Independently from us \citet{delayed_syntax} have employed delayed gradient updates in syntax NMT setting, where the sequences are much longer due to the syntax annotation and delayed updates are necessary because video RAM is limited.\footnote{This work was published on 01.05.2018.}

Independently from us \citet{local_sgd} have developed their own local optimizer solution as an alternative to increasing mini-batch sizes.\footnote{This work was published on 22.08.2018}

\section{Conclusion and Future work}
We show that we can increase speed and maintain convergence rate for very large mini-batch asynchronous SGD by carefully adjusting momentum and applying warmup and cooldown strategies. While we have demonstrated our methods on GPUs, they are hardware agnostic and can be applied to neural network training on any multi-device hardware such as TPUs or Xeon Phis. We were able to achieve end-to-end training on multiple tasks a lot faster than the baseline systems. For our Romanian-English model, we train nearly 3X faster than the commonly used baseline and 1.5X faster over a specifically optimised baseline. When experimenting with English-German we are able to train our model 1.3X faster than the baseline model, achieving practically the same BLEU score and much better model cross-entropy.

In the future we would like to apply local optimizers in distributed setting where the communication latency between local and remote devices varies significantly we could use local optimizers to synchronize remote models less often.

\section*{Acknowledgments}
We thank Adam Lopez, Sameer Bansal and Naomi Saphra for their help and comments on the paper. We thank our reviewers for their comments and suggestions. Nikolay Bogoychev was funded by an Amazon faculty research award to Adam Lopez.

\bibliography{naaclhlt2018}
\bibliographystyle{acl_natbib}

\end{document}